# Approximately Optimal Monitoring of Plan Preconditions

Craig Boutilier
Department of Computer Science
University of Toronto
Toronto, ON M5S 3H5
cebly@cs.toronto.edu

## Abstract

Monitoring plan preconditions can allow for replanning when a precondition fails, generally far in advance of the point in the plan where the precondition is relevant. However, monitoring is generally costly, and some precondition failures have a very small impact on plan quality. We formulate a model for optimal precondition monitoring, using partially-observable Markov decisions processes, and describe methods for solving this model effectively, though approximately. Specifically, we show that the single-precondition monitoring problem is generally tractable, and the multiple-precondition monitoring policies can be effectively approximated using single-precondition solutions.

## 1 Introduction

Uncertainty in planning problems is often handled by modeling the problem deterministically—enabling classical planning techniques to be used—but using methods for execution monitoring and replanning to handle situations that arise when the plan fails (e.g., when a precondition at some point fails to hold). Two extreme approaches can be adopted: The first requires monitoring all preconditions required by future actions in the plan (once they are established); when one fails replanning is invoked. Refinements of this scheme are, of course, possible.[1] The second, and much more common, approach simply monitors the current state and should an unanticipated state be reached replanning in invoked [8, 13, 7, 16, 2]. Variants of this scheme include the use of universal plans [14], essentially performing replanning in advance.

While both of these approaches have a certain appeal, they each have some rather serious drawbacks. The first approach allows one to anticipate precondition failures well in advance and replan as soon as one notices that the current plan will not work. However, it does not allow for the fact that precondition failure may be temporary or intermittent (e.g., a traffic jam may render a plan infeasible, but perhaps should not be taken into account if it occurs on a route that will not be reached for several hours). Even worse it does not factor in the cost of monitoring. Generally, precondition monitoring is not cost-free (e.g., tasking an agent to monitor a route, or obtaining information from a Web source, has some cost). As a consequence, the value of monitoring a precondition from the time it is established until the time it is used may not be worth the cost (e.g., knowing about traffic many hours in advance is not likely to be much more useful than learning of it just before reaching the desired route, assuming a reasonable alternative route can be found at the later point in time). The second approach suffers from the opposite problem: though monitoring costs play no role, one cannot anticipate failures in advance. This generally means that the best repaired plan is not as good as one constructed when the failure is known in advance (e.g., if the traffic jam is not discovered until one is *in it*, the best repaired plan is likely of poor quality compared to one that avoided the jammed route entirely).

The decision of whether to monitor a plan precondition, and when to monitor it, involves balancing the cost of monitoring and the value of monitoring information. Specifically, the value of a monitoring report at any point in time depends on the odds that a report could change the plan, and the value of the best plan should that report not be received. As such, we can formulate the problem of monitoring as a *partially-observable Markov decision process* (POMDP). To do so requires that we have available the following information prior to plan execution time:

- the probability that preconditions may fail
- the cost of attempting to execute a plan action when its precondition has failed
- the value of the best alternative plan at any point during plan execution (i.e., given that we abandon the current plan at that point)
- a model of monitoring processes and their accuracy (e.g., the probability that a precondition is reported to be OK when it has in fact failed).

We assume that such information can be obtained or estimated, and discuss these assumptions further below.

---

[1] For example, Veloso [16] monitors selected conditions for *opportunities* to construct better plans.

Unfortunately, though optimal monitoring can be formulated as a POMDP, for any nontrivial plan, the size of the required POMDP takes it far from the realm of practical solution. Plans of only three steps (and three preconditions) severely tax state-of-the-art algorithms. For this reason we propose a class of heuristic techniques for solving the optimal monitoring problem. These methods involve solving the monitoring problem for *individual* preconditions, then constructing an (online) monitoring policy based on these component solutions. Though the individual problems also involve solving POMDPs, these remain very small and tractable. Thus the construction of monitoring policies for plans involving hundreds of steps is rendered feasible using our heuristic methods. We demonstrate empirically on a small selection of problems that the solution quality of our techniques is generally quite good.

The main contributions of this work are twofold. We first provide a decision-theoretic model of plan monitoring. This model makes clear the role that value of information plays in optimal plan monitoring and the sequential nature of the decision problem. Neither of these characteristics is present in the existing plan-monitoring literature. The second contribution is a tractable class of methods for solving the plan monitoring problem. Though these methods are heuristic and we currently explore their quality empirically, we expect that they should yield to theoretical error analysis. These algorithms make the abstract monitoring problem computationally manageable, thus making our decision-theoretic model a practical alternative to standard classical plan-monitoring methods.

## 2  The Plan Monitoring Problem

Classical planning techniques have advanced to the point where large planning problems involving hundreds of actions in sophisticated domains (e.g., logistics, process planning) can be solved effectively. However, these models invariably assume away uncertainty, modeling problems deterministically. Even though this modeling assumption may be reasonable, uncertainty (e.g., in action effects, or exogenous events) must be dealt with when it impacts the ability to execute the plan. Plan and execution monitoring and replanning are often used in the regard.

The simplest monitoring model involves monitoring the established preconditions of every action at each point in time until that action is executed. If a precondition has failed (e.g., due to unanticipated exogenous events) then we replan from the current point in the plan subject to the observed constraints. Unfortunately, despite offering optimal object-level performance, this model may be too costly to implement. If precondition monitoring has some cost, the expected benefit in terms of improved object-level (repaired) plan quality may not outweigh the monitoring costs. Furthermore, if the monitoring reports are subject to error (e.g., unreliable Web sources, or faulty sensors), this approach is not satisfactory. A more refined decision-theoretic model is required, one where the probability of precondition failure and the cost of precondition monitoring are balanced against the expected improvement in plan quality offered by timely information about the status of that precondition. This allows one to optimally decide *if* and *when* to monitor a given precondition. In this section, we formulate a specific version of the plan monitoring problem and describe how it can be solved optimally when cast as a POMDP.

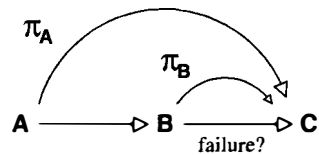

Figure 1: A Sample Scenario

It is important to note that a planning problem where precondition failure and monitoring are both possible can be directly posed and solved as a POMDP. Specifically, assuming the availability of the information required for optimal plan monitoring (e.g., precondition failure probabilities and monitoring accuracy models), a POMDP for the *planning* problem—as opposed to the *plan monitoring* problem—can be formulated readily. Forcing the problem into a classical framework, even with good execution monitoring and replanning strategies, will generally lead to suboptimal behavior. Despite this, breaking up the problem by constructing a classical plan together with a monitoring policy can prove fruitful for several reasons. Most importantly, the POMDP formulation will generally prove impossible to solve for all but the most trivial problems. As such, the classical model can be viewed as a way of approximating the solution of the underlying POMDP. This isn't to say that other ways of approximating the POMDP's solution would not also be appropriate; and there has been little if any research on how to directly form a deterministic relaxation of a POMDP or bound the quality of the classical plans so formed. But this form of solution has the advantage of relying on widely-used (and often very efficient) classical planning technology. Apart from this, the monitoring model we propose can be used with any classical plan, regardless of how it was constructed (e.g., it may simply be a plan constructed by a human expert). Since the classical view of plans as sequences of actions is often very natural in many domains, plan monitoring remains an important problem to be tackled using decision-theoretic techniques.

### 2.1  A Motivating Example

To illustrate the types of tradeoffs that must be addressed in plan monitoring, consider the following simple route planning example, illustrated in Figure 1. The best (e.g., lowest cost) plan $\pi$ to reach goal location $C$ from initial location $A$ traverses the bottom-most links through location $B$. We informally say that action $A$ moves from $A$ to $B$, thus the plan consists of two actions, $A$ and $B$. If the link $B \to C$ fails (e.g., become impassable), the best plan from $A$ involves an alternative route $\pi_A$, and similarly the best plan from $B$ is $\pi_B$. We can monitor this link at any point in time for a certain cost, say, just before execution of $A$ or $B$. If we learn that a link has failed, we can adopt the best alternative plan for the point at which failure was discovered.

Intuitively, one should monitor the link $B \to C$ at some point if the *expected value of information* (EVOI) obtained by monitoring outweighs the cost of obtaining that information. For example, suppose we monitor $B \to C$ just prior to execution of action $B$. If we learn of link failure, the best "repaired" plan $\pi_B$ has value $v(\pi_B)$. If we had not learned of this failure, we would have continued execution of the original plan $\pi$, with a failure occurring when we try to execute $B$. We assume that this failure will be repaired when it occurs, giving us an plan with value $v(\pi_{fail})$. Therefore, the value of learning of link failure at point $B$ is given by $v(\pi_B) - v(\pi_{fail})$.[2] EVOI is given by

$$EVOI(B) = \Pr(B) \cdot (v(\pi_B) - v(\pi_{fail}))$$

where $\Pr(B)$ is the probability of link failure occurring by "time" $B$. Monitoring just prior to $B$ is then worthwhile if the EVOI is greater than the monitoring cost.

We will develop a model below that makes precisely these tradeoffs. However, there are a number of subtleties that must be dealt with to provide an accurate account. First, as presented above, we have assumed that *perfect* information is available, when, in fact, monitoring information is likely to be error-prone or uninformative. To deal with this we assume the existence of "sensor models" that describe the probability that a monitoring report is faulty in various ways. This also requires that we maintain a *belief state* describing the probability of precondition failures based on previous monitoring reports; we will seldom know of failure with certainty. We must also account for the sequential nature of the problem. Applying this reasoning to monitoring prior to action $A$ might suggest that monitoring at that point is also useful. But, in fact, this decision depends on whether one should monitor at point $B$. If it is worthwhile monitoring at $B$, it may not be worthwhile to *additionally* monitor at $A$. Intuitively, if $v(\pi_B)$ is close to $v(\pi_A)$, then monitoring at $A$ is probably not worthwhile, while if $v(\pi_A)$ is much greater, it probably is. The sequential nature of the problem demands a dynamic programming formulation of monitoring policy construction.

### 2.2 Modeling Assumptions

We assume that we have a deterministic planning problem, which has been solved with a classical plan $\pi$; this is a sequence of (deterministic) actions $\langle a_1, a_2, \cdots, a_n \rangle$. We somewhat loosely use the term *time t* to refer to the point just prior to the execution of $a_t$ (thus time ranges from 1 to $n+1$). The preconditions for action $a_t$ all hold at time $t$ in normal plan execution. For simplicity we assume each action has a single precondition; thus we use the terms "monitoring an action" and "monitoring a precondition" interchangeably.

In general, preconditions should only be considered for monitoring at certain points in the plan. The *monitoring interval* for $a_t$ is the interval between the establishment of the precondition of $a_t$ and time $t$.[3] There is no point considering the monitoring of $a_t$ outside of this interval. Again for expository purposes, we assume that relevant preconditions have been established prior to the execution of the plan; that is, no action $a_t$ establishes the (monitored) precondition for a future action $a_{t+k}$. This is merely to keep notation to a minimum—the techniques that follow make no important use of this fact.

We assume a plan *value* or cost model: for any alternative plan $\pi'$, we know the value $v(\pi')$ of that plan. This may, for instance, simply ascribe higher value to plans with lower total action cost. Apart from knowing $v(\pi)$ for the original plan $\pi$, we also assume that we can determine by planning, or estimate by some other means, the value of the *best alternative plans* at each time $t$. Specifically, if we know that the precondition for $a_k \in \pi$ has failed, the best alternative plan $\pi_j$ is known—or at least some estimate of its value $v(\pi_j)$—for each $1 \leq j \leq k$. So if we abandon $\pi$ at time $j$ because some future precondition has failed, the best alternative has a known value. We also know the value of attempting to execute an action in the original plan when its precondition does not hold, and subsequently implementing the best repaired plan. We dub this plan $\pi_k^f$. (If the truth of preconditions can't be unknown before an action is attempted, we simply need to set $\pi_k^f = \pi_k$.)

Though actions are modeled as though they have deterministic effects, certain exogenous events can occur that destroy established preconditions. In order to construct optimal monitoring strategies, we must have some model of the likelihood of precondition failure. We adopt a general model of exogenous events, using a *spontaneous transition model* $T(s'|s)$, where $s$ and $s'$ denote arbitrary system states. The quantity $T(s'|s)$ denotes the probability that the system state $s$ will transition to state $s'$ due to the occurrence of some exogenous event (or events).[4] From the point of view of plan monitoring, we are interested only in precondition failure, so the required state space $\mathcal{S}$ simply consists of all truth assignments to plan preconditions.[5] To ease the modeling burden, we adopt the assumption of *precondition failure independence*, requiring that the probability of one precondition changing its state is independent of any other. Thus we can model transitions using $n$ $2 \times 2$ transition matrices, where $T^j$ denotes the dynamics of the $j$th precondition. Specifically, $T^j(\neg p|p)$ denotes the probability of precondition failure, while $T^j(p|\neg p)$ denotes the probability of spontaneous precondition repair (e.g., clearing of a traffic jam). Dealing with complex events that induce correlations in the failures of different preconditions can be modeled in probabilistic STRIPS notation [9, 10], dynamic Bayes nets [6, 3], or other representations. We need not move to full $2^n \times 2^n$ transition matrices.

A set of *monitoring actions* $\mathcal{M}$ is assumed, each action pro-viding information about a particular precondition and having a fixed cost. We use a *sensor model* of monitoring action $m$ to determine its influence on our *degree of belief* in the precondition (i.e., estimate of the probability that the precondition holds). For any action $m$ that monitors precondition $p$, we assume: (a) a finite set of possible observation values $Z_m$ that can be returned by $m$; and (b) a stochastic sensor model that specifies, for each $z \in Z_m$, both $\Pr(z|p)$ and $\Pr(z|\neg p)$. To keep the presentation simple, we assume that observations are restricted to "true" and "false," and the sensor model dictates the probability of false positive and false negative reports on a precondition. This type of model derives from the standard model used in partially observable MDPs [1, 15, 11]. Monitoring actions have no influence on the underlying state of the world (though this could easily be incorporated, as it is in the general POMDP formulation). Rather they influence only our assessment of a precondition's truth. We assume a fixed monitoring cost $c_t$ for the monitoring of each action $a_t$.[6]

---

[2] We require that $v(\pi_B) \geq v(\pi_{fail})$ since one could always ignore failure if it is discovered.

[3] A precondition $p$ for $a_t$ is *established* by action $a_j$, $j < t$, if $a_j$ makes $p$ true and no intervening action $a_k$, $j < k < t$ affects $p$. If no such $a_j$ makes $p$ true, it is established by the initial state. More generally, we can define the monitoring interval for each precondition $p_i(a_t)$.

[4] The *stationarity assumption*, where the transition probabilities are fixed at each time step, can easily be relaxed.

[5] We assume preconditions are boolean variables for ease of presentation; however, our model can easily be extended to deal with discrete failure *modes*.

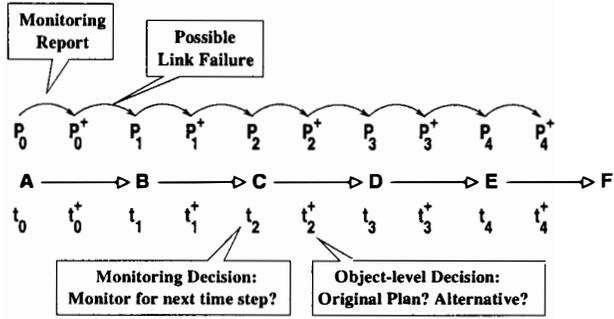

Figure 2: Sequencing of Decision Stages and Belief Update

The sequence of object-level and monitoring decisions, and the points at which one's belief state will be updated, are illustrated graphically in Figure 2 (other models of sequencing are possible however). At any point in time $1 \leq t \leq n$, assuming the agent is still able to execute the original plan $\pi$, the agent makes two decisions in sequence. First it can request the monitoring of any action $a_k$ for any $k \geq t$. The observation reports obtained are used to update its current belief state $P_t$ to obtain a new distribution over preconditions $P_t^+$. Precondition independence means that this distribution can be factored into $n$ distributions over individual preconditions. Given this updated belief state, the agent now makes an object-level action choice: it can either *continue* with the original plan $\pi$ and execute action $a_t$, or it can *abandon* the plan and execute the alternative plan $\pi_t$. If the original plan is abandoned due to the possibility of link failure, then no future decisions need to be made: we assume that $\pi_t$ can be executed without difficulty to provide value $v(\pi_t)$. If $a_t$ is executed, any precondition's status can change: the system dynamics determine the probability with which this occurs. The agent can update its belief state from $P_t^+$ to $P_{t+1}$ reflecting the possible changes. As a consequence, there are two decision stages for each plan step: a *monitoring stage* with a decision at time $t$, where one decides which preconditions to monitor; and an *action stage* decision at time $t^+$, the decision regarding which object-level action to execute (i.e., to continue or abandon the plan).

### 2.3 A POMDP Formulation

Optimal monitoring decisions can be determined by casting the problem as a POMDP and solving using standard dynamic programming methods [15, 11].

A POMDP can be viewed as a fully observable Markov decision process whose state space consists of probability distribution over underlying system states, or *belief states*. While the set of belief states is continuous (with dimension $|\mathcal{S}| - 1$), Sondik [15] showed that the $k$-stage-to-go value functions $V^k$ of a finite-horizon POMDP is piecewise linear and convex (p.w.l.c.) and thus can be represented finitely using a collection of linear functions over belief space, or $\alpha$-*vectors*. Specifically, given such a collection $\aleph^k$ of $|\mathcal{S}|$-dimensional $\alpha$-vectors, $V^k(b) = \max\{b \cdot \alpha : \alpha \in \aleph^k\}$.

The sequence of value functions $V^k$ (or more accurately, their representation as sets $\aleph^k$) can be computed by dynamic programming. We describe the basic intuitions using Monahan's [12] algorithm since it is conceptually straightforward. However, we tailor the presentation to our monitoring problem. We assume an $n$-step plan monitoring problem with $n$ action stages and $n$ monitoring stages. The final decision occurs at time $n+$. A decision to continue with the last step of the original plan or to abandon the plan is made based on the belief state $b^{n+}$. The value of *abandoning* the plan is $v(\pi_n)$ regardless of the true state: hence the $Q$-*function* $Q_{aban}^{n+}$ for action *aban*, where $Q_{aban}^{n+}(b)$ is the value of *aban* at belief state $b$, can be represented by the constant vector $\alpha_{aban}$ with entries $v(\pi_n)$. The value of *cont* (i.e., attempting to *continue* the original plan) is simply $v(\pi)$ if the final precondition holds and $v(\pi_n^f)$ if it does not. Thus $Q_{cont}^{n+}(b) = b \cdot \alpha_{cont}$, where $\alpha_{cont}$ is the vector with entries $v(\pi)$ for each state where the precondition holds, and $v(\pi_n^f)$ where it doesn't. The value function $V^{n+}$ is thus representable by $\aleph^{n+}$ comprising these two vectors. It is clear that $V^{n+}$ is p.w.l.c. We note that each vector is associated with a specific action: if

$$b \cdot \alpha_{cont} \geq b \cdot \alpha_{aban}$$

then the optimal choice is *cont*, otherwise it is *aban*. More generally, as we see below, vectors denote the value of executing a complete course of action (or *conditional plan*) over the remainder of the problem horizon.

Given $\aleph^{t+}$ for the $t$th action stage, we can compute $\aleph^t$ for the preceding monitoring stage as follows. One can choose to monitor any subset of the remaining $t$ preconditions; thus a monitoring action refers to some *collection* of individual monitoring actions. Any such compound monitoring action chosen involving $k \leq t$ preconditions gives rise to $2^k$ possible observations (2 observations for each observed condition). Since a different course of action can be pursued after each distinct observation, we define the set *observation strategies* $OS_m^t$ for time $t$ and monitoring action $m$ to be the set of mappings $\sigma : Z_m \to \aleph^{t+}$ that associate a subsequent $\alpha$-vector with each possible observation (note that each vector has a conditional plan associated with it). The value of executing $m$ together with $\sigma$ is again a linear function of the belief state. Specifically, for each state $s$ the probability $\Pr(z|s)$ of any observation $z$ is fixed, and the $Q$-value of $\sigma$ at $s$ is:

$$Q(\sigma, s) = c(m) + \sum_z \{\Pr(z|s)\sigma(z)_s\}$$

---
[6]More realistic models that distinguish the costs of continuing vs. intermittent monitoring could also be adopted.

The Q-value of $\sigma$ can thus be represented by the vector $\alpha^\sigma$ with $s$th component $Q(\sigma, s)$; and $Q^t(\sigma, b) = b \cdot \alpha^\sigma$ for any belief state $b$. The best monitoring action and observation strategy at belief state $b$ is simply that which has maximum expected value at $b$; thus the value of optimal monitoring can be represented by the collection of $\alpha$-vectors induced by the strategies in $\{OS_m^t : m \in \mathcal{M}\}$. For each $\alpha$-vector in this set we record the appropriate monitoring action: if $\alpha^\sigma$ corresponds to $\sigma \in OS_m^t$, we associate $m$ with $\alpha^\sigma$.[7]

Finally, given the value function $\aleph^{t+1}$ for the $t+1$st monitoring stage, it is a simple matter to compute the value function $\aleph^{t+}$ for the $t$th action stage. The decision at time $t+$ is, again, whether to continue the original plan $\pi$ or abandon it (executing $\pi_t$). If the agent persists with $\pi$ for one additional step, value is given almost directly by $V^{t+1}$: once the specific step of $\pi$ is executed, the agent can act optimally by selecting the subsequent course of action dictated by $V^{t+1}$. Each course of action corresponds to some $\alpha \in \aleph^{t+1}$. The value of continuing with $\pi$ at a specific state $s$ followed by implementing $\alpha$ is given by:

$$Q^{t+}(\alpha, s) = \sum_{s' \in \mathcal{S}} T(s'|s)\alpha_{s'}$$

for any $s$ where the precondition for $a_t$ holds. At all other $s$, $Q^{t+}(\alpha, s) = v(\pi_t^f)$. The value of continuing with $\pi$ can therefore be represented by:

$$Q_{cont}^{t+} = \{\langle Q^{t+}(\alpha, s) : s \in \mathcal{S}\rangle : \alpha \in \aleph^t\}$$

With each such vector we associate the action *cont*. If $\pi$ is abandoned, the value obtained is the constant $c(\pi_t)$ (independent of the state $s$). Thus

$$\aleph^{t+} = \{\langle c(\pi_t), ...c(\pi_t)\rangle\} \bigcup Q_{cont}^{t+}$$

where the action *aban* is associated with the constant vector.

Apart from the division into action and monitoring stages, this algorithm is essentially that proposed by Monahan with one exception. The collections of $\alpha$-vectors defined above typically contain many *dominated* vectors that do not maximize value at any belief state. Monahan's algorithm involves an additional pruning phase where dominated vectors are removed at each stage before moving to the next stage. This provides tremendous computational benefit.

---

[7] We note that not monitoring any precondition corresponds to choosing the empty subset of conditions above. The $Q$-value of this action is simply identical to the value function $V^{t+}$, thus the set $\aleph^{t+}$ can be copied directly into $\aleph^t$.

Other algorithms, including *linear support* [5] and *Witness* [4] proceed by directly identifying only (or primarily) non-dominated vectors and thus require little or no pruning, and tend to be more efficient still. Our results in Section 4 are all based on the Witness algorithm.

Given a collection of $\aleph$-sets, implementation of the monitoring and execution policy requires that the agent maintain and update its belief state $b$ over time. At each (monitoring or action) stage $k$, $b$ is applied to each vector $\alpha \in \aleph^k$ to determine $b \cdot \alpha$, and the action associated with the maximizing vector is executed. Actions are either monitoring decisions for the remaining preconditions, or "continue" decisions. At action stages, the single *aban*-vector has constant value, so the *cont*-vectors need only be searched until one better than the sole *aban*-vector is found.

While this model is conceptually appealing, it is computationally intractable for all but the most trivial plan monitoring problems. For plans involving $n$ preconditions, there are $2^n$ states, as many as $2^n$ monitoring actions, and up to $2^n$ observations. Present (exact) POMDP algorithms can at best deal with problems involving a thousand states and are highly sensitive to the number of actions and observations. The solution of the plan monitoring POMDP can be well beyond the reach of state-of-the-art algorithms like Witness for *three* step plans. Clearly, some problem decomposition and approximation is required if the decision-theoretic approach is to be practical.

## 3 Heuristic Monitoring

In this section we consider two alternative models for solving the monitoring problem that are vastly superior to the full POMDP formulation computationally. Intuitively, for a planning problem with $n$ stages, we solve $n$ independent monitoring problems, one for each precondition (recall that we assume a single precondition per action). The solutions to these individual problems are then combined *online*. In particular, at a given decision stage, whether action or monitoring, we have access to the value functions and policy decisions for the *individual* problems at that stage, as well as the current belief state. These are used to determine an appropriate choice of action for the original monitoring problem at that point.

### 3.1 Solving Single-Failure POMDPs

For each precondition (or action) $a_t$ in an $n$-stage plan, we consider the problem of optimally monitoring $a_t$ over the interval $[1, t]$ under the assumption that this is the *only* precondition that can fail. This is a $t$-stage POMDP with action and monitoring stages as above. The key difference is that decisions are based on belief in the state of that precondition alone. As such the corresponding POMDP has only two states, one monitoring action, and two observations. This $t$th *single-failure monitoring problem* is therefore generally very easily solved. The solution of each of $n$ single-failure problems provides us with a $t$-stage value function for the $t$th problem, this value function consisting of $t$ sets of $\alpha$-vectors, $\aleph_t^1, \cdots, \aleph_t^t$. Thus the value functions for all $n$ problems can be represented in $O(l \cdot n^2)$ space, where $l$ is a bound on the size of any $\aleph$-set.

## 3.2 Naive Policy Combination

Assume that the $n$ single-failure monitoring POMDPs for an $n$-stage planning problem have been solved. At any given monitoring stage $t$, the individual policies for each action $a_k$ ($t \leq k \leq n$) will each require that their action either be monitored or not. At each action stage $t+$, the polices for each $a_k$ will either suggest that the original plan be abandoned or continued.

Our *Naive Policy Combination (NPC)* algorithm works as follows. The agent maintains a factored belief state $\langle b_1, \cdots, b_n \rangle$ over the $n$ individual preconditions (since these are independent). At each monitoring stage $t$, the individual value functions $\aleph_k^t$ ($t \leq k \leq n$) are applied to the $b_k$, and the monitoring decision for each $a_k$ made on this basis. Thus the actual monitoring action $m = \langle m_t, \cdots, m_n \rangle$ executed has $m_k$ assigned to "monitor" iff monitoring is optimal for the $k$th subproblem. At each action stage $t+$, the individual value functions $\aleph_k^{t+}$ ($t \leq k \leq n$) are applied to the $b_k$, and the object-level action decision for each $a_k$ determined. The action *cont* is executed if *each* of the individual policies suggest continuing. The action *aban* is executed if *any* of the individual policies suggest abandonment.

## 3.3 Value Function Adjustment

In the NPC strategy described above, if any monitoring problem requires that the plan be abandoned, then abandonment is (globally) optimal. However, continuing may not necessarily be optimal. Consider for instance that each precondition may be probable enough that the "abandonment threshold" for the *individual* problem is not reached. However, when the probability of failure *as a whole* is considered, abandonment may, in fact, be appropriate.

We can perform some simple adjustments to the individual value functions to attain a more accurate estimate of the value of continuing a plan. Consider the decision at action stage $t+$, with individual value functions $\aleph_k^{t+}$ ($t \leq k \leq n$) and belief state $b_k$ ($1 \leq k \leq n$). The value $V_n^{t+}(b_n)$ is an accurate reflection of value assuming that all preconditions $a_j$ ($j < n$) preceding $a_n$ are OK. Specifically, the course of action associated with any $\alpha \in \aleph_n^{t+}$ does in fact have value $b_n \cdot \alpha$. The value $V_{n-1}^{t+}(b_{n-1})$ is also accurate under the assumption that the preceding conditions $a_j$ ($j < n-1$) are OK as well as condition $a_n$. However, if we wish to account for the fact that $a_n$ may have failed, we can adjust the value function to reflect the fact that the impact of $a_n$ failing is captured by the value $V_n^{t+}(b_n)$ just calculated.

This adjustment can be effected by noting that for any $\alpha \in \aleph_{n-1}^{t+}$, its $i$th component $\alpha_i$ is given by $p_i^\alpha v(\pi) + (1 - p_i^\alpha) x$, where $p_i^\alpha$ denotes the probability of successfully reaching and executing action $a_{n-1}$ under the conditional plan reflected by $\alpha$ at state $s_i$, and $x$ is some indeterminate quantity (reflecting average value of plan abandonment/failure). Once we execute $a_{n-1}$, however, the value $v(\pi)$ is not assured for action $a_n$ may fail, or be abandoned, etc. To approximate the influence of this possibility, we can replace $v(\pi)$ in the above expression with the expected value of the $n$th monitoring problem, $V_n^{t+}(b_n)$; we replace each component $\alpha_i$ of $\alpha$ with

$$\alpha_i - p_i^\alpha (v(\pi) - V_n^{t+}(b_n))$$

This *value-adjusted* estimate offers a better picture of the overall value of executing the conditional plan associated with $\alpha$, taking into account the influence of the later precondition. We then take $V_{n-1}^{t+}(b_{n-1})$ to be computed w.r.t. the adjusted $\alpha$-vectors $\widehat{\aleph}_{n-1}^{t+}$.

Our *Value-Adjusted Policy Combination (VAPC)* algorithm works as follows. Monitoring decisions are made precisely as in NPC. The action decision at stage $t+$ is made by working backwards from stage $n$ to stage $t$. We define $\widehat{\aleph}_k^{t+}$ for each $t \leq k < n$ to be the set of value-adjusted vectors using $\widehat{V}_{k+1}^{t+}(b_{k+1})$ (i.e., each $\alpha \in \aleph_k^{t+}$ is replaced by its value-adjusted counterpart using $\widehat{V}_{k+1}^{t+}(b_{k+1})$ as the substitute for $v(\pi)$). $\widehat{V}^{t+}$ is in turn defined as the value function induced by the value-adjusted $\aleph$-set $\widehat{\aleph}_{k+1}^{t+}$. This recursion is grounded at stage $n$ where $\widehat{\aleph}_n^{t+} = \aleph_n^{t+}$. Algorithmically, this process can be implemented efficiently. Starting at stage $n$, the continue/abandon decision is made for $a_n$. If the decision is *aban*, a global abandon decision is made and we terminate. Otherwise, we move to stage $n-1$, computing the adjusted set $\widehat{\aleph}_{n-1}^{t+}$ using $V_n^{t+}(b_n)$ (note that $V_n^{t+}(b_n)$ is computed as a by-product of the decision for $a_n$). A decision to continue or abandon is then computed for $a_{n-1}$ using $\widehat{\aleph}_{n-1}^{t+}$. If *aban* is chosen, again we abandon the plan, otherwise we move to stage $n-2$, adjusting its $\alpha$-vectors using the (already-computed) value $\widehat{V}_{n-1}^{t+}(b_{n-1})$. This is repeated back to $a_t$, terminating whenever one action calls for abandonment, or when $a_t$ is reached with all actions calling for continuation.

The probabilities $p_i^\alpha$ can can be computed easily during the dynamic programming solution of the individual monitoring problem for $a_k$. For each $\alpha$-vector $\langle v_1, v_2 \rangle$ at any stage, we compute a corresponding probability vector $\langle p_1^\alpha, p_2^\alpha \rangle$. (there are two states, one denoting $a_k$'s precondition OK, and one its failure). At stage $n$, these probabilities are either 0 or 1 (recall they are a function of the state, not belief state). At any earlier stage, they are calculated as a function of the probabilities associated with the following stage, in exactly the same way that the values $v_i$ for the $\alpha$-vector are computed. This adds minimal computation time to dynamic programming and requires only that we store an extra collection of vectors: a probability vector for every $\alpha$-vector.

The additional online adjustment phase of VAPC makes online policy combination slightly more complex than NPC: VAPC requires roughly twice the time to come to a decision at any stage. Both, however, are linear in the sum of the sizes of the value functions being used. If the vector sets for each subproblem are bounded in size, then this is linear in the plan size (number of stages). Thus both methods are efficient in their online computations.

## 4 Empirical Results

In this section we describe empirical results suggesting that the computation time required to solve monitoring problems using our approximation technique is negligible when compared to the full POMDP model, and that it scales very well to plans involving many hundreds of steps. We also provide evidence that the solution quality is generally quite good. Our ability to do so is limited however by the fact that computing optimal solutions for all but the most trivial problems is a practical impossibility. The Witness algorithm is used to solve all POMDPs (both the full POMDP and the single-failure POMDPs for each problem).[8]

We begin with a simple three-stage problem. This problem has characteristics that make it relatively "easy" to solve: precondition failure has small probability and no precondition can become OK once it has failed. The failure and abandonment costs do not impose severe penalties, so the value function has few components.[9] This full POMDP was still very difficult to solve, requiring 9608 seconds (2.7 hrs) of CPU time for Witness. Despite this the largest value function (at the first monitoring stage) had only eight vectors (each precondition was monitored in some of the corresponding actions, though not every combined monitoring action was part of the optimal policy). In contrast, the approximation algorithm produced the collection of component value functions in 5.78 seconds (5.86 seconds if adjustment probabilities are computed, 2.71 seconds if Monahan's algorithm is used rather than Witness).[10] The largest $\aleph$-set had only 4 vectors.

On other three-stage problems, we were unable to get Witness to run to completion in a reasonable time on the full POMDP. For instance, in one example with a more complex value function, Witness was terminated after 76035 seconds (over 21 hours of CPU time) with an agenda (see [4] for details) of over 10000 vectors (with indications that the agenda was still growing). In contrast, the approximation method required only 5.06 seconds (5.08 if adjustment probabilities are computed).[11]

The scaling of the approximation algorithms is illustrated in Figure 3, where solution time is plotted as a function of problem size for a series of related problems. In this sequence, the scaling appears to be nearly linear, though in fact this is largely due to the fact that the value functions tend to simplify as the horizon grows. In general, we expect scaling to be quadratic in problem size. The (rather slow) quadratic growth is evident in Figures 4 and 5, where the solution times for the component single-failure monitoring problems in a 400-step plan are shown, as well as the cumulative solution time (thus this 400-step monitoring problem is solved in about 4 hours).[12]

---

[8] Algorithms are implemented in Matlab and run under Linux on a 550MHz PIII architecture with 512Mb of memory.

[9] Specifically, each of the three preconditions had a 0.01 chance of failing at each stage, and observation of each precondition has a 0.1 false negative rate and a 0.3 false positive rate. Successful plan value is 20; alternative plans have value 12, 8 and 4, respectively, at steps 1 through 3; plan failure values are 10, 5, and 2 (hence plan failure does not impose great cost); observation costs are 0.5, 0.5 and 0.7 for preconditions 1 through 3, respectively.

[10] Monahan's algorithms tends to work better than Witness if the value functions are very compact.

[11] This problem is similar to the one above but with a higher precondition failure rate of 0.05, more accurate observations (0.2 false positive rate), smaller observation costs (0.3), and greater cost due to plan failure (i.e., greater difference between successful and failed plan values).

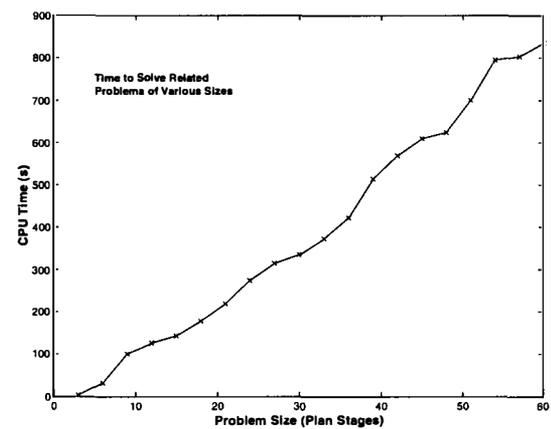

Figure 3: Solution Time as a Function of Problem Size

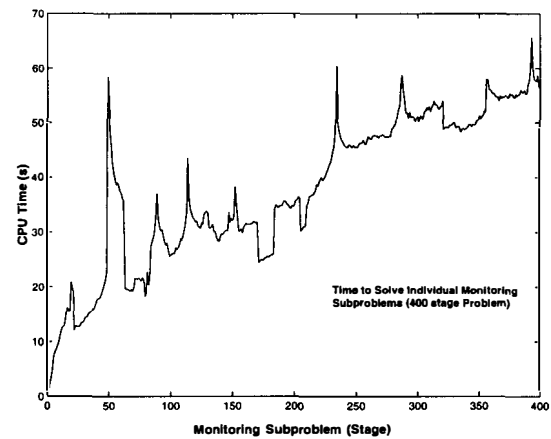

Figure 4: Solution Time for Component Subproblems (400 Stage Problem)

The efficiency gains of this approach cost very little in terms of solution quality. We compare solution quality of our approximations to the optimal solution for the three-stage problem we were able to solve optimally (see details above). This comparison is made by comparing the expected value of the policy induced by NPC and VAPC with the optimal value, at a number of different belief states. We sampled 1331 belief states uniformly distributed over belief space: for each of the three variables, each degree of belief between 0 and 1 was sampled at intervals of 0.1. Over these 1331 states, the average relative error in decision quality for NPC was 0.049, and for VAPC, 0.047; thus on average both strategies give rise to policies whose value is within 5% of optimal. The maximum relative error at any belief state is 0.166 for NPC and 0.142 for VAPC. The sacrifice in decision quality seems a small price to pay for the vast difference in computational effort.

---

[12] Note that the full POMDP has $2^{400}$ states, $2^{400}$ monitoring actions and up to $2^{400}$ observations for certain monitoring actions.

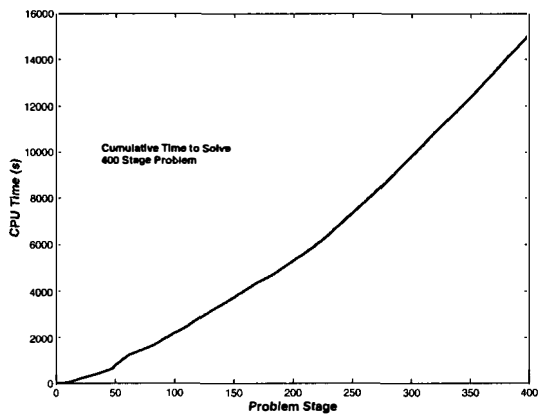

Figure 5: Cumulative Solution Time for Subproblems (400 Stage Problem)

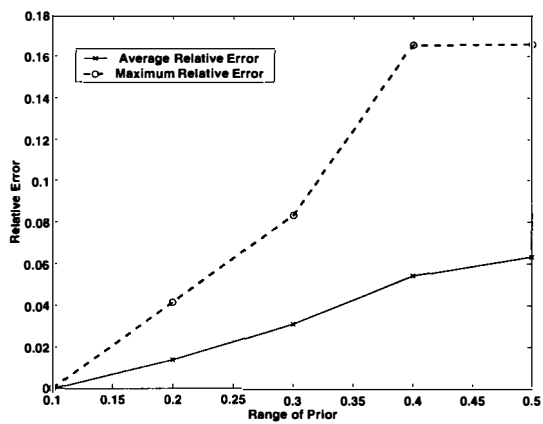

Figure 7: Solution Quality over "Low Prior" Belief States

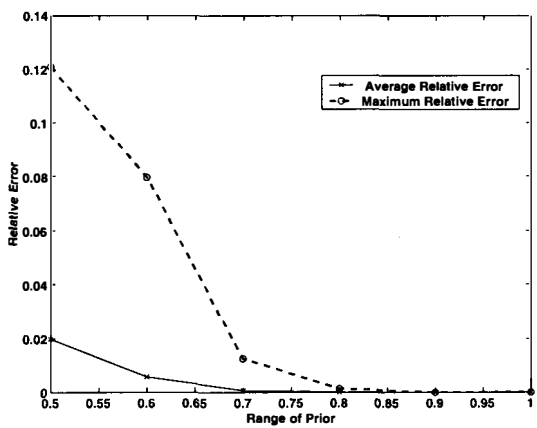

Figure 6: Solution Quality over "High Prior" Belief States

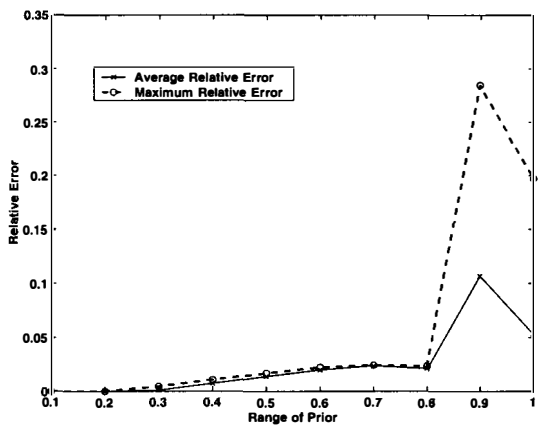

Figure 8: Solution Quality of NPC vs. VAPC

More interesting is the fact that the decision quality tends to vary significantly in different parts of belief space. We plot the relative value error for "high prior" belief states in Figure 6 (here we only show NPC). At each point $p$ (e.g., 0.7), we show the relative errors over all sampled belief states where each precondition is restricted to have prior probability between $p$ and 1 in increments of 0.1 (e.g., at 0.7 we see the error ranging over belief states, for each variable, in the set $\{0.7, 0.8, 0.9, 1.0\}$). We see here that the relative error for our approximation technique tends to decrease significantly at belief states with high prior precondition probability. At .9 and above, all decisions are in fact optimal. At .8 and above, average error is about 0.1 per cent. This is important because plans involving such preconditions are likely to be invoked only when the preconditions are reasonably likely to hold.

A similar plot for "low prior" states is shown in Figure 7, where relative error for belief states ranging from $p$ down to 0 (again in 0.1 increments) is plotted. Again we note that the error tends to be most pronounced in the intermediate belief states.

Finally, we compare the relative value attained by NPC and VAPC on a slightly larger 5-stage problem.[13] The solution of this problem requires 16.04 seconds (16.09 when adjustment probabilities are computed). Figure 8 illustrates the relative improvement of VAPC over NPC on a number of belief state ranges. Each point $p$ shows the average and maximum relative improvement of VAPC over the 243 belief states where each of the 5 variables has probability in $\{p - 0.1, p - 0.05, p\}$. In this problem, the individual value functions tend to be reasonably sensitive in the choice of whether to continue of abandon in the neighborhood $[0.8, 0.9]$. VAPC offers considerable advantage of NPC areas around this point, with average improvement in decision quality of nearly 11% in this range and maximum improvement of 28.5%.

---

[13]Problem parameters: 0.05 probability of precondition failure and 0.1 chance of precondition repair; observations have a 0.1 false negative and a 0.2 false positive rate; successful plan value is 40; alternative plan values are 25, 18, 12, 7, and 6; failure values are 12, 11, 7, 5, and 2; observation costs range from 0.3 to 0.5.

## 5 Concluding Remarks

We have described a decision-theoretic model for optimal plan monitoring that takes into account monitoring costs, the probability of precondition failure, and the value of alternative plans. While this model is conceptually appealing, it is wildly intractable, leading us to develop approximation methods that scale very well with plan size and seem to make small sacrifices in decision quality. This approach makes decision-theoretic monitoring practical for complex planning and monitoring problems.

There are a number of simple improvements that can be made to this approach. One involves scaling to large planning problems through the use of *critical points*. One can identify a *subset* of a plan's actions to monitor—rather than monitoring all $n$ actions—by considering the difference in the value of the alternative plans $\pi_t$ at various points. If $v(\pi_t)$ is not much less than $v(\pi_{t+1})$ it may be reasonable to simply ignore action $a_t$ in one's monitoring problem. By judicious selection of such critical points (i.e., points in the plan such that the cost of abandoning the plan once committing to them is very high), the number of stages and preconditions one needs in a monitoring problem can be reduced.

The extension of this model to handle correlated precondition failures is critical for many applications. Dealing with correlations should prove to be fairly easy, using, say, Bayesian networks to represent existing independence in the transition model and the belief state. The same basic approach to decomposing the POMDP into individual monitoring problems is still applicable, though the value adjustment phase in the VAPC technique will require modification. Other directions for future research include developing formal error bounds for this approach, the incorporation of more sophisticated cost and value models for the underlying planning domain, and extending the model to deal with partially-ordered plans.

**Acknowledgements** Many thanks to Pascal Poupart and Manuela Veloso for their helpful discussion, as well as to the reviewers for their comments. This research was supported by IRIS Phase 3 Project BAC, NSERC Research Grant OGP0121843, and the DARPA Co-ABS program (through Stanford University contract F30602-98-C-0214).